\documentclass[letterpaper, 10 pt, conference]{ieeeconf}

\usepackage{shellesc}

\IEEEoverridecommandlockouts %
\usepackage{amsmath,amsfonts,amssymb}
\usepackage{mathrsfs}
\DeclareMathOperator{\atan}{atan}
\usepackage{bbold} %
\usepackage{bbding} %
\newcommand{\medstar}{\text{\FiveStarOpen}} %

\usepackage{subcaption}
\usepackage{tabularx}
\usepackage{multirow}
\usepackage{siunitx}
\usepackage{xfrac}

\usepackage[export]{adjustbox}

\usepackage{graphicx}
\graphicspath{{figures/}}

\usepackage{caption}
\usepackage{xargs} %

\usepackage[style=ieee, backend=biber, maxbibnames=3, mincitenames=1, maxcitenames=2, url=false, doi=false, isbn=false, citestyle=numeric-comp, natbib=true, eprint=true, bibencoding=utf8]{biblatex}

\usepackage{bm} %
\usepackage{lipsum}

\usepackage{pgfplots}
\usepackage{pgfgantt}
\usepackage{pdflscape}
\pgfplotsset{compat=newest}
\pgfplotsset{plot coordinates/math parser=false}

\usepackage{tikz}
\usetikzlibrary{decorations.markings, positioning}

\usepackage{tikzscale}

\newlength\figH
\newlength\figW

\title{\LARGE \bf Linear Differential Games for Cooperative Behavior Planning of Autonomous Vehicles Using Mixed-Integer Programming}
\author{Tobias Kessler$^{1*}$, Klemens Esterle$^{1*}$ and Alois Knoll$^{2}$%
 	\thanks{$^{*}$These authors contributed equally to this work.}%
	\thanks{$^{1}$Tobias Kessler and Klemens Esterle are with fortiss GmbH, Research Institute of the Free State of Bavaria, Munich, Germany, surname@fortiss.org}%
	\thanks{$^{2}$Alois Knoll is with Robotics, Artificial Intelligence and Real-time Systems, Technische Universit\"{a}t M\"{u}nchen, Munich, Germany}%
}

\begin{document}

\newcommand\copyrighttext{%
	\scriptsize \textcolor{blue}{\textcopyright 2020 IEEE. Personal use of this material is permitted.  Permission from IEEE must be obtained for all other uses, in any current or future media, including reprinting/republishing this material for advertising or promotional purposes, creating new collective works, for resale or redistribution to servers or lists, or reuse of any copyrighted component of this work in other works}}
\newcommand\copyrightnotice{%
	\begin{tikzpicture}[remember picture,overlay]
	\node[anchor=north,yshift=-7.5pt] at (current page.north) {\fbox{\parbox{\dimexpr\textwidth-\fboxsep-\fboxrule\relax}{\copyrighttext}}};
	\end{tikzpicture}%
}

\maketitle
\copyrightnotice
\thispagestyle{empty}
\pagestyle{empty}

\global\csname @topnum\endcsname 0
\global\csname @botnum\endcsname 0

\newcommand{\figurename}{Fig. }

\newcommand {\vect} {\boldsymbol}
\newcommand {\matr} {\boldsymbol}

\newcommand{\state} {\vect{x}}
\newcommand{\stateSpace} {\vect{\mathcal{X}}}
\newcommand{\beliefstate} {\vect{b}}
\newcommand{\contr} {\vect{u}}
\newcommand{\contrSpace} {\vect{\mathcal{U}}}
\newcommand{\meas} {\vect{y}}
\newcommand{\procNoise}{\vect{w}}

\newcommand {\cov}  {\matr{\Sigma}}

\newcommand{\stateNoDelta}{\hat\state}
\newcommand{\contrNoDelta}{\hat\contr}
\newcommand{\procNoiseNoDelta}{\hat\procNoise}

\newcommand{\abc}[2][\empty]{%
	\ifthenelse{\equal{#1}{\empty}}
	{no opt, mand.: \textbf{#2}}
	{opt: \textbf{#1}, mand.: \textbf{#2}}
}

\newcommand {\noiseu} {\procNoise}
\newcommand {\covu} {\matr{\Sigma_{\noiseu,}}}

\newcommand {\noiseuNoDelta} {\procNoiseNoDelta}
\newcommand {\covuNoDelta} {\matr{\Sigma_{\noiseuNoDelta,}}}

\newcommand {\defnoiseu}[1][\empty]{
	\ifthenelse{\equal{#1}{\empty}}
	{\noiseu\sim N(0,\covu)}
	{\noiseu_{#1}\sim N(0,\covu_{#1})}
}

\newcommand {\defnoiseuNoDelta}[1][\empty]{
	\ifthenelse{\equal{#1}{\empty}}
	{\noiseuNoDelta\sim N(0,\covuNoDelta)}
	{\noiseuNoDelta_{#1}\sim N(0,\covuNoDelta_{#1})}
}

\newcommand {\covm} {\matr{R}}
\newcommand {\noisem} {\vect{\nu}}
\newcommand {\defnoisem}[1][\empty]{
	\ifthenelse{\equal{#1}{\empty}}
	{\noisem\sim N(0,\covm)}
	{\noisem_{#1}\sim N(0,\covm_{#1})}
}

\newcommand{\stateB}{\vect{\xi}}
\newcommand{\contrB}{\vect{\nu}}
\newcommand{\procNoiseB}{\vect{\omega}}

\newcommand{\AB}{\mathcal{A}}
\newcommand{\BB}{\mathcal{B}}
\newcommand{\WB}{\mathcal{W}}
\newcommand{\costStateB}{\mathcal{Q}}
\newcommand{\costContrB}{\mathcal{R}}
\newcommand{\covStatesB}{\mathcal{S}_{\state}}
\newcommand{\covProcNoiseB}{\mathcal{S}_{\procNoise}}

\newcommand{\FB}{\mathcal{F}}

\newcommand{\cct}{\vect{t}}
\newcommand{\ccsval}{s}
\newcommand{\ccT}{\matr{T}}
\newcommand{\ccsvec}{\vect{s}}

\newcommand{\costState}{\matr{Q}}
\newcommand{\costContr}{\matr{R}}
\newcommand{\feedbackMatrix}{\matr{K}}
\newcommand{\cost}{J}

\newcommand{\stateConstraintMatrix}{\matr{C}}
\newcommand{\stateConstraintVector}{\vect{c}}

\newcommand{\stateConstraintFunc}{c}

\newcommand{\contrConstraintMatrix}{\matr{D}}
\newcommand{\contrConstraintVector}{\vect{d}}

\newcommand{\contrConstraintFunc}{d}

\newcommand{\stateRef}{\state^{*}}
\newcommand{\contrRef}{\contr^{*}}

\newcommand{\stateDelta}{\Delta\state}
\newcommand{\contrDelta}{\Delta\contr}

\newcommand {\Comment}[1]{\textcolor{blue}{#1}}

\newcommand {\partialder}[4][\bigg]{\frac{\partial #2}{\partial #3}#1|_{#4}}
\newcommand {\partialdernoarg}[3][\bigg]{\frac{\partial #2}{\partial #3}#1}

\newcommand{\nat}{\mathbb{N}}
\newcommand{\real}{\mathbb{R}}
\newcommand{\compl}{\mathbb{C}}

\newcommand{\norm}[1]{\left\| #1 \right\|}

\newcommand{\half}{\frac{1}{2}}

\newcommand{\parenth}[1]{ \left( #1 \right) }
\newcommand{\bracket}[1]{ \left[ #1 \right] }
\newcommand{\accolade}[1]{ \left\{ #1 \right\} }
\newcommand{\pardevS}[2]{ \delta_{#1} f(#2) }
\newcommand{\pardevF}[2]{ \frac{\partial #1}{\partial #2} }

\newcommand{\vecii}[2]{\begin{pmatrix} #1 \\ #2 \end{pmatrix}}
\newcommand{\veciii}[3]{\begin{pmatrix}  #1 \\ #2 \\ #3	\end{pmatrix} }
\newcommand{\veciv}[4]{\begin{pmatrix}  #1 \\ #2 \\ #3 \\ #4	\end{pmatrix}}

\newcommand{\matii}[4]{\left[ \begin{array}[h]{cc} #1 & #2 \\ #3 & #4 \end{array} \right]}
\newcommand{\matiii}[9]{\left[ \begin{array}[h]{ccc} #1 & #2 & #3 \\ #4 & #5 & #6 \\ #7 & #8 & #9	\end{array} \right]}

\newcommand{\transp}{^{\intercal}}
\newcommand{\Reg}{$^{\textregistered}$}
\newcommand{\reg}{$^{\textregistered}$ }
\newcommand{\Tm}{\texttrademark}
\newcommand{\tm}{\texttrademark~}
\newcommand {\bsl} {$\backslash$}

\newtheorem{theorem}{Theorem}[section]
\newtheorem{lemma}[theorem]{Lemma}
\newtheorem{corollary}[theorem]{Corollary}
\newtheorem{remark}[theorem]{Remark}
\newtheorem{definition}[theorem]{Definition}
\newtheorem{equat}[theorem]{Equation}
\newtheorem{example}[theorem]{Example}
\newcommand{\insertfigure}[4]{ %
	\begin{figure}[htbp]
		\begin{center}
			\includegraphics[width=#4\textwidth]{#1}
		\end{center}
		\vspace{-0.4cm}
		\caption{#2}
		\label{#3}
	\end{figure}
}

\newcommand{\refFigure}[1]{\figurename \ref{#1}}
\newcommand{\refChapter}[1]{Chapter \ref{#1}}
\newcommand{\refSection}[1]{Section \ref{#1}}
\newcommand{\refParagraph}[1]{Paragraph \ref{#1}}
\newcommand{\refEquation}[1]{(\ref{#1})} %
\newcommand{\refTable}[1]{Table \ref{#1}}
\newcommand{\refAlgorithm}[1]{Algorithm \ref{#1}}

\newcommand{\rigidTransform}[2]
{
	${}^{#2}\!\mathbf{H}_{#1}$
}

\newcommand{\code}[1]
{\texttt{#1}}

\newcommand{\comment}[1]{\marginpar{\raggedright \noindent \footnotesize {\sl #1} }}

\newcommand{\clearemptydoublepage}{%
	\ifthenelse{\boolean{@twoside}}{\newpage{\pagestyle{empty}\cleardoublepage}}%
	{\clearpage}}

\newcommand{\etAl}{\emph{et al.}\mbox{ }}

\makeatletter
\def\ignorecitefornumbering#1{%
	\begingroup
	\@fileswfalse
	#1%
	\endgroup
}
\makeatother

\newcommand{\pxy}{p_\medstar}
\newcommand{\vxy}{v_\medstar}
\newcommand{\axy}{a_\medstar}
\newcommand{\uxy}{u_\medstar}
\newcommand{\xy}{\square_\medstar}
\newcommand{\deltat}{\Delta t}
\newcommand{\x}{p_x}
\newcommand{\y}{p_y}
\newcommand{\vx}{v_x}
\newcommand{\vy}{v_y}
\newcommand{\ax}{a_x}
\newcommand{\ay}{a_y}
\newcommand{\jx}{j_x}
\newcommand{\jy}{j_y}
\newcommand{\ux}{u_x}
\newcommand{\uy}{u_y}
\newcommand{\xfrontlb}{\underline{f_x}}
\newcommand{\yfrontlb}{\underline{f_y}}
\newcommand{\xfrontub}{\overline{f_x}}
\newcommand{\yfrontub}{\overline{f_y}}
\newcommand{\xyfrontlb}{\underline{f_\medstar}}
\newcommand{\xyfrontub}{\overline{f_\medstar}}
\newcommand{\squarefrontlb}{\underline{f_{\medstar}}}
\newcommand{\squarefrontub}{\overline{f_{\medstar}}}
\newcommand{\xfront}{f_{x}}
\newcommand{\yfront}{f_{y}}
\newcommand{\xyfront}{f_{\medstar}}
\newcommand{\xref}{p_{x,ref}}
\newcommand{\yref}{p_{y,ref}}
\newcommand{\vxref}{v_{x,ref}}
\newcommand{\vyref}{v_{y,ref}}
\newcommand{\orientation}{\theta}
\newcommand{\orientationlb}{\underline{\orientation}}
\newcommand{\orientationub}{\overline{\orientation}}
\newcommand{\region}{\rho} %
\newcommand{\possibleregion}{\varrho^r}
\newcommand{\umaxx}{\overline{\ux^r}}
\newcommand{\umaxy}{\overline{\uy^r}}
\newcommand{\uminx}{\underline{\ux^r}}
\newcommand{\uminy}{\underline{\uy^r}}
\newcommand{\uminxy}{\underline{\uxy^r}}
\newcommand{\umaxxy}{\overline{\uxy^r}}
\newcommand{\amaxx}{\overline{\ax^r}}
\newcommand{\amaxy}{\overline{\ay^r}}
\newcommand{\aminx}{\underline{\ax^r}}
\newcommand{\aminy}{\underline{\ay^r}}
\newcommand{\aminxy}{\underline{\axy^r}}
\newcommand{\amaxxy}{\overline{\axy^r}}
\newcommand{\noregionchange}{\Psi}
\newcommand{\noregionchangexpos}{\noregionchange_x^{+}}
\newcommand{\noregionchangexneg}{\noregionchange_x^{-}}
\newcommand{\noregionchangeypos}{\noregionchange_y^{+}}
\newcommand{\noregionchangeyneg}{\noregionchange_y^{-}}
\newcommand{\vregchange}{V}%
\newcommand{\vmin}{\underline{v}}
\newcommand{\vmax}{\overline{v}}
\newcommand{\amin}{\underline{a}}
\newcommand{\amax}{\overline{a}}
\newcommand{\umin}{\underline{u}}
\newcommand{\umax}{\overline{u}}
\newcommand{\fractionparametersone}{\alpha}
\newcommand{\fractionparameterstwo}{\beta}
\newcommand{\fractionparametersthree}{\gamma}
\newcommand{\fractionparametersfour}{\delta}
\newcommand{\polysinub}{\overline{\mathscr{P}_{\sin}^r}}
\newcommand{\polycosub}{\overline{\mathscr{P}_{\cos}^r}}
\newcommand{\polysinlb}{\underline{\mathscr{P}_{\sin}^r}}
\newcommand{\polycoslb}{\underline{\mathscr{P}_{\cos}^r}}
\newcommand{\polycosubone}{\overline{\chi_C}}
\newcommand{\polycosubtwo}{\overline{\psi_C}}
\newcommand{\polycosubthree}{\overline{\omega_C}}
\newcommand{\polysinubone}{\overline{\chi_S}}
\newcommand{\polysinubtwo}{\overline{\psi_S}}
\newcommand{\polysinubthree}{\overline{\omega_S}}
\newcommand{\polycoslbone}{\underline{\chi_C}}
\newcommand{\polycoslbtwo}{\underline{\psi_C}}
\newcommand{\polycoslbthree}{\underline{\omega_C}}
\newcommand{\polysinlbone}{\underline{\chi_S}}
\newcommand{\polysinlbtwo}{\underline{\psi_S}}
\newcommand{\polysinlbthree}{\underline{\omega_S}}
\newcommand{\polykappaub}{\overline{\mathscr{P}_{\kappa}^r}}
\newcommand{\polykappalb}{\underline{\mathscr{P}_{\kappa}^r}}
\newcommand{\polykappa}{\mathscr{P}_{\kappa}^r}
\newcommand{\polykappaminone}{\underline{\chi_\kappa}}
\newcommand{\polykappamintwo}{\underline{\psi_\kappa}}
\newcommand{\polykappaminthree}{\underline{\omega_\kappa}}
\newcommand{\polykappamaxone}{\overline{\chi_\kappa}}
\newcommand{\polykappamaxtwo}{\overline{\psi_\kappa}}
\newcommand{\polykappamaxthree}{\overline{\omega_\kappa}}
\newcommand{\notwithinenv}{e}
\newcommand{\envpolys}{\Lambda}
\newcommand{\envpolyline}{\lambda}
\newcommand{\envpolylinesegment}{l}
\newcommand{\envpolylinesegmentpointone}{a}
\newcommand{\envpolylinesegmentpointtwo}{b}
\newcommand{\deltacc}{o_p}
\newcommand{\deltaccall}{o_\medstar}
\newcommand{\deltaccfrontlblb}{o_{\underline{f}\underline{f}}}
\newcommand{\deltaccfrontlbub}{o_{\underline{f}\overline{f}}}
\newcommand{\deltaccfrontublb}{o_{\overline{f}\underline{f}}}
\newcommand{\deltaccfrontubub}{o_{\overline{f}\overline{f}}}
\newcommand{\obstacle}{\mathscr{o}}
\newcommand{\obstacleset}{\mathscr{O}}
\newcommand{\setofregions}{\mathscr{R}}
\newcommand{\timeintervall}{\mathscr{K}}
\renewcommand{\t}{k}
\newcommand{\StateMatrix}{\bm{A}}
\newcommand{\InputMatrix}{\bm{B}}
\newcommand{\setOfAgents}{\mathscr{A}}
\newcommand{\Agent}{A}
\newcommand{\carToCarCollision}{\alpha}
\newcommand{\agentSafetyDist}{D}
\newcommand{\slack}{\xi}

\begin{abstract}
Cooperatively planning for multiple agents has been proposed as a promising method for strategic and motion planning for automated vehicles.
By taking into account the intent of every agent, the ego agent can incorporate future interactions with human-driven vehicles into its planning.
The problem is often formulated as a multi-agent game and solved using iterative algorithms operating on a discretized action or state space.
Even if converging to a Nash equilibrium, the result will often be only sub-optimal.
In this paper, we define a linear differential game for a set of interacting agents and solve it to optimality using mixed-integer programming. 
A disjunctive formulation of the orientation allows us to formulate linear constraints to prevent agent-to-agent collision while preserving the non-holonomic motion properties of the vehicle model. 
Soft constraints account for prediction errors. 
We then define a joint cost function, where a cooperation factor can adapt between altruistic, cooperative, and egoistic behavior.
We study the influence of the cooperation factor to solve scenarios, where interaction between the agents is necessary to solve them successfully. 
The approach is then evaluated in a racing scenario, where we show the applicability of the formulation in a closed-loop receding horizon replanning fashion.
By accounting for inaccuracies in the cooperative assumption and the actual behavior, we can indeed successfully plan an optimal control strategy interacting closely with other agents.
\end{abstract}

\IEEEpeerreviewmaketitle

\section{Introduction}

\begin{figure}[t]
	\centering
	\includegraphics[width=0.57\linewidth]{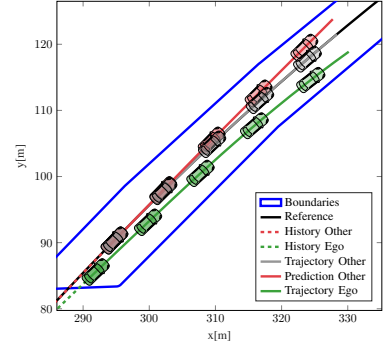}
	\caption{Example of a multi-agent planning scenario. The green ego agent wants to overtake the red agent but is unaware of its exact future motion (gray). Both agents intend to track the same reference line and must stay inside the road boundaries (blue). Dark-to-light colors depict progressing time.}
	\label{fig:intro}
\end{figure}

When sharing the road with human drivers, autonomous vehicles will have to cooperate with other vehicles to fit safely into nowadays traffic scenarios while still asserting their own goals.
In dense traffic, where space is often limited, such as highway overtaking or merging, the reactions of others must be anticipated, and a maneuver-based prediction will perform poorly. 
Instead, a planner must model the uncertain interaction with the other traffic participants by planning a joint action for the ego vehicle and the surrounding vehicles.
\refFigure{fig:intro} depicts an exemplary scenario. 

A variety of concepts have been proposed to cope with these challenges, with game-theoretic approaches being capable of modeling the interaction between multiple agents elegantly \cite{Schwarting2018}.
However, they often require a discretization of the action or the state space, yielding an approximation of the optimal solution, which are usually work well for a limited set of scenarios.
Furthermore, these algorithms often rely on a random sampling of the solution space or lack guarantees of convergence and thus pose open questions towards certification of such systems.
Optimal control theory, on the other hand, provides deterministic solution algorithms converging to an optimum and has been successfully applied to safety-critical systems.

In this work, we propose an approach how an autonomous agent can safely interact with other agents while still achieving its own goals.
For this, we formulate the multi-agent planning problem as a differential game and solve it using mixed-integer quadratic programming (MIQP) with an off-the-shelf solver.
Specifically, we contribute a linear differential game formulation featuring
\begin{itemize}
	\item a set of linear constraints as inter-agent collision check,
	\item a leveraged joint cost function for collaborative planning, and
	\item a methodology to handle inaccurate models of other agents using soft constraints.
\end{itemize}

We first examine the impact of cooperation in a symmetric negotiation example.
We then demonstrate the contributions in a competitive race track example as this scenario poses high demands on the accurate modeling of vehicle kinematics, handling of vehicle collisions, and interactive behavior.

\begin{table*}[t]
	\centering
	\vspace{0.15cm}
	\caption{Comparison of different solution approaches for multi-agent planning using the abbreviations mixed-integer quadratic programming (MIQP), mixed-integer linear programming (MILP), Monte Carlo tree search (MCTS), dynamic programming (DP).}
	\scriptsize
	\begin{tabular}{l|l|l|l|l|l}
Source & Interactiveness & Action Space & State Space & Solution Method & Globalization Strategy \\
\hline 
\ignorecitefornumbering{\citet{Kessler2019}} & Global cooperative costs & discrete & cont. & MILP & W.r.t. sampled motion tree\\
\ignorecitefornumbering{\citet{Lenz2016}} & Multi-agent dynamic game, non-zero-sum & discrete & cont. & MCTS & None \\
\ignorecitefornumbering{\citet{Bahram2016}} & Two-player dynamic game & discrete & cont. & Alpha-beta pruning & None\\	
\ignorecitefornumbering{\citet{Liniger2020}} & Multi-agent dynamic game, non-zero-sum & cont. & cont. & DP & Solving Nash equilibrium \\
\ignorecitefornumbering{\citet{Schwarting2019}} & Multi-agent dynamic game, non-zero-sum & cont. & cont. & Iterative DP & Solving iterative Nash equilibrium \\
\ignorecitefornumbering{\citet{Fabiani2020a}} & Potential game & mixed & mixed & MIQP & Solving iterative Nash equilibrium\\
\hline
\ignorecitefornumbering{\citet{Eilbrecht2017}} & Global cooperative costs, shared plans & cont. & cont. & MIQP & Iterative conflict resolution\\
\ignorecitefornumbering{\citet{Manzinger2018}} & Centralized planning, shared plans & cont. & discrete & Reachability & Auction-based conflict resolution\\
\ignorecitefornumbering{\citet{Frese2011}} &  Global cooperative costs, centralized & cont. & cont.  & MILP & Inherently \\
\hline
\ignorecitefornumbering{\citet{Esterle2020}} & None & cont. & cont. & MIQP & Inherently \\
\end{tabular}%
	\label{tab:model_comparism}
\end{table*}

\section{Related Work}
We follow the distinction of \citet{Ulbrich2015b} dividing cooperative driving into explicit inter-vehicle communication and cooperation in the form of collaboration.
In \refTable{tab:model_comparism} we compare some approaches with comparable problem formulations or similar solution methods.
For a more comprehensive overview we refer to \citet{Schwarting2018}.

Game-theoretic formulations can be employed to model collaboration. A decision tree often realizes the game-theoretic setting.
\citet{Kessler2019} use motion primitives to generate a motion tree and use mixed-integer linear programming (MILP) to solve for the optimal orchestration. While it is applicable in arbitrary environments, the action space discretization can exclude the optimal solution, yielding difficulties when applying to dense traffic scenarios. They account for unknown cost functions of other agents by updating the costs based on observations.
Monte Carlo tree search can be used to plan collaborative behavior, effectively solving a multi-agent, non-zero-sum dynamic game \cite{Lenz2016}. A cooperation factor serves as a tuning parameter in the ego agents' cost function. The formulation is highly flexible and can incorporate any transition function for modeling the environment. However, the discretized action space yields similar problems to \cite{Kessler2019}.
An extensive-form game is formulated in \cite{Bahram2016}, where the other traffic participants are modeled as part of the environment. While the framework is highly flexible and has proven to work in a real car under real-time requirements, the approach does not ensure convergence to an optimal solution.
A two-player dynamic, non-zero-sum game is formulated as a bimatrix game in \cite{Liniger2020}, which allows for an efficient calculation of the Nash equilibrium.
\citet{Schwarting2019} use iterative dynamic programming to solve a multi-agent dynamic, non-zero-sum game. They explicitly model partial observability of the intention of others. The proposed believe-space variant of the iterative Linear Quadratic Gaussian (iLQG) algorithm can be executed in real-time.
Exact costs and dynamics of other agents are assumed to be known, and the algorithm converges to a potentially sub-optimal Nash equilibrium.
Multi-vehicle driving as a potential game is formulated in \cite{Fabiani2020a} and solved using MIQP. The potential function allows the authors to compute a $\epsilon$-mixed-integer Nash equilibrium. However, the discrete lateral action and state space complicate applying this approach in reality.

For explicit communication between vehicles, the planning problem effectively simplifies, as accounting for the unknown intentions of other agents becomes irrelevant. 
\citet{Eilbrecht2017} formulate a two-layered approach of iterative conflict resolution using a cooperative cost function. The underlying behavior of each agent is generated through an optimal control problem using MIQP for each agent while ensuring obstacle avoidance to the (known) plan of the other agents. However, their approach is only valid for straight driving on straight roads, and the iterative conflict resolution does not offer any guarantees to converge to a global optimum.
\citet{Manzinger2018} use reachability analysis to compute conflicting space-time cells, which might be occupied by multiple vehicles. An auction algorithm then solves for those conflicts. Their approach requires a discretization of the state space.
Various approaches exist to minimize a cooperative cost function based on the premise that the cost functions of the other agents are known. \citet{Frese2011} formulate a multi-agent optimal control problem and solve it using MILP. However, their approach is limited to straight driving on straight roads and yields a lot of invalid solutions otherwise. Our previous work \cite{Esterle2020} solves those issues but was not defined for a multi-agent context.

\section{Problem Formulation and Assumptions}

We aim to find a control strategy in an environment with multiple agents, subject to the following set of assumptions. 
Firstly, we only control one agent but indirectly assume perfect observation of the dynamic state of other agents, as well as their goals.
Secondly, we assume a perfect perception of the map and our localization within that.
And thirdly, that other agents do not behave destructively, for example, do not aim for collisions.
The problem then consists of
\begin{itemize}
	\item multiple agents with continuous actions,
	\item a scene consisting of all individual states, road geometry, and obstacle information,
	\item each agent tries to achieve its own goals, without adversarial behavior, and
	\item perfect observation of states at $t_0$.
\end{itemize}
We formulate this as a multi-agent, non-zero-sum, differential game \cite{Steven2004} with
\begin{itemize}
	\item a set of agents $\setOfAgents$ with continuous actions, and
	\item a joint, non-zero-sum, cost function.
\end{itemize}

The strategy is executed in a receding horizon fashion.
We discretize the time horizon of one iteration into $N$ steps with a time interval $\deltat$. We denote the discrete time step by $\t$ and the interval of $N$ steps by $\timeintervall$. The goal is thus to find a sequence of actions for the ego vehicle that minimizes its costs while avoiding collisions with the environment.

\section{Linear Dynamic Game using MIQP}
\label{sec:miqp_model}

In this section, we formulate the differential game and then solve it using MIQP. The linear constraint formulation yields a linear differential game.
Each agent in the optimization problem satisfies the constraints from our previous work \cite{Esterle2020}, which we discussed in \refSection{sec:preliminaries}.
Also, we introduce linear constraints for agent-to-agent collision checking in \refSection{sec:agenttoagentcollision}, and a joint cost function in \refSection{sec:costfunction}.

\subsection{Preliminaries}
\label{sec:preliminaries}

The model in this paper builds upon our previously proposed model. 
In the following, we will highlight the contributions of our previous work; for a detailed model formulation we refer to \citet{Esterle2020}.

\paragraph{Linear Model}%
We model the vehicle as a third-order point-mass system in a global Cartesian frame with positions $\x(\t)$, $\y(\t)$, velocities $\vy(\t)$, $\vy(\t)$, and accelerations $\ax(\t)$, $\ay(\t)$ as states.
Jerk in both directions $\jx(\t)$ and $\jy(\t)$ are the inputs of the model.
This model is subject to a set of constraints for collision avoidance and non-holonomy.

Although the vehicle's orientation $\orientation$ is not part of the state space, we need it for a sufficient collision check in the Cartesian coordinates within the optimization problem.
We define \textit{regions} in the $(\vx, \vy)$ plane, allowing us to obtain an approximation of the orientation based on linear inequalities only dependent on $v_x$ and $v_y$.
Note that computing the true orientation $\orientation = \atan(\vy/\vx)$ would introduce non-linearity.
A binary decision variable $\region$ defines in which region the vehicle is at each time step. It is set by constraints based on the upper and lower linear bounds of each region.
We impose region-dependent limits on acceleration and jerk.

We approximate the collision shape of the vehicle using circles. 
We over-approximate the front axle collision shape because the true front axle position $(\xfront, \yfront)$ is not directly known from the linear model.
Computing the upper and lower bounds for sine and cosine functions of the orientation leads to upper $\xfrontub$, $\yfrontub$ and lower bounds $\xfrontlb$, $\yfrontlb$ for the $x$ and $y$ position of the front axle.
\refFigure{fig:agent_to_agent_collision} illustrates the approximation.
For obtaining the lower and upper bounds for sine and cosine functions of the true orientation, we fit linear polynomials on $\vx$ and $\vy$.

\paragraph{Modeling the Non-Holonomics}
To ensure non-holonomics of a vehicle for a point-mass system, we constrain the curvature, that is highly non-linear and thus cannot be expressed as a linear constraint in MIQP. 
We thus model the non-holonomics as linear constraints on the acceleration in the $y$-direction, for which we provide the theory.
For that, the curvature is approximated using a linear combination of $\vx$, $\vy$, $\ax$, and is solved for $\ay$.

\paragraph{Environment and Obstacle Collision Avoidance}
The polygonal representation of the environment is deflated with the radius of the collision circles. Non-convex environment polygons are split into several convex sub-polygons. We enforce the vehicle to be in at least one of these convex sub-polygons. This way, the vehicle-to-environment-polygon collision check becomes a point-to-polygon check for each circle used to approximate the vehicle.

We ensure the vehicle not to collide with an arbitrary number of static or dynamic convex obstacle polygons, such as obtained from a prediction. 
However, our previous work does not implement a vehicle-to-vehicle collision check, which we contribute in this work.

\subsection{Multi-Agent Collision Constraints}
\label{sec:agenttoagentcollision}
We approximate the shape of each vehicle to formulate an agent-to-agent collision constraint.
In our linear model, the actual orientation of the vehicle is not available, but with the region-based formulation, we can compute a lower- and upper-bounding rectangle for the center of the front axle.

\begin{figure}[tb]
	\vspace{0.15cm}
	\footnotesize
	\centering
	\def\svgwidth{0.65\columnwidth}
	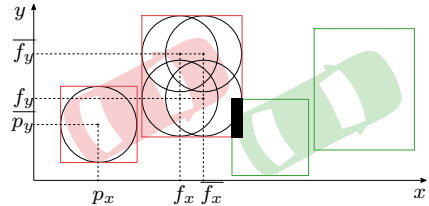
	\caption{Approximation of the vehicle shape to formulate the agent-to-agent collision check based on the rectangles of the respective agents. The black area indicates a collision.}
	\label{fig:agent_to_agent_collision}
	\vspace{-0.15cm}
\end{figure}

In the following, the superscript $\square^i$ refers to the respective variable of the agent $i$.
For collision avoidance, we approximate the vehicle shape by circles with radius $R^i$, one around the rear axle center, and four for the front axle approximation.
To avoid agent-to-agent collisions, we choose to over-approximate these circles with axis-aligned squares again, as sketched in \refFigure{fig:agent_to_agent_collision}.
A better approximation of the circles, for example, with two rectangles, yields more constraints and binary variables, lowering the runtime without providing a huge benefit.
We formulate four sets of constraints, the first prevents collisions between the rear parts of two agents, the second prevents collisions between the rear part of the first and the front part of the second agent, the third vice versa, and the fourth prevents collisions between the front parts of both agents for each pair of agents.

The rear part-to-rear part collision constraint of two agents $\Agent^i$ and $\Agent^j$ is based on the following logical formula.
We define the sum of both radii $R^{i+j} := R^i + R^j$.
A collision occurs at one time step $\t$ if and only if
\begin{align}
\x^i(\t) \geq \x^j(\t) - R^{i+j} &\land \x^i(\t) \leq \x^j(\t) + R^{i+j} \nonumber\\
\land ~ \y^i(\t) \geq \y^j(\t) - R^{i+j} &\land \y^i(\t) \leq \y^j(\t) + R^{i+j}.
\label{eq:rear_rear_collision}
\end{align}
Intuitively, \refEquation{eq:rear_rear_collision} states that a collision occurs if both the absolute distance in $x$-direction $|\x^i(\t) - \x^j(\t)|$ and $y$-direction $|\y^i(\t) - \y^j(\t)|$ is smaller $R^{i+j}$.
Logical negation yields that two agents do not collide at time step $\t$ if and only if
\begin{align}
\x^i(\t) \leq \x^j(\t) - R^{i+j} &\lor \x^i(\t) \geq \x^j(\t) + R^{i+j} \nonumber\\
\lor \y^i(\t) \leq \y^j(\t) - R^{i+j} &\lor \y^i(\t) \geq \y^j(\t) + R^{i+j}. 
\label{eq:rear_rear_no_collision}
\end{align}
We formulate this as linear constraints using a set of four decision variables $\carToCarCollision^{ij}_\square$, one for each inequality and an appropriately chosen big constant $M$.
\begin{subequations}
\begin{align}
\x^i(\t) &\leq \x^j(\t) - R^{i+j} + M \carToCarCollision^{ij}_1(\t) \label{eq:rear_rear_no_collision_1}\\ 
\x^i(\t) &\geq \x^j(\t) + R^{i+j} - M \carToCarCollision^{ij}_2(\t)\\
\y^i(\t) &\leq \y^j(\t) - R^{i+j} + M \carToCarCollision^{ij}_3(\t)\\ 
\y^i(\t) &\geq \y^j(\t) + R^{i+j} - M \carToCarCollision^{ij}_4(\t) \label{eq:rear_rear_no_collision_4}\\
3 &\geq \sum_{a=1}^4  \carToCarCollision^{ij}_a(\t) ~ \forall ~ \t \in \timeintervall. \label{eq:rear_rear_no_collision_coupling}
\end{align}
\label{eq:rear_rear_no_collision_constaint}
\end{subequations}
\refEquation{eq:rear_rear_no_collision_coupling} represents the logical formulas \refEquation{eq:rear_rear_no_collision} by coupling the four constraints \refEquation{eq:rear_rear_no_collision_1} - \refEquation{eq:rear_rear_no_collision_4} and makes sure no more than three are active, and hence no rear part-to-rear part collision occurs.

As we aim to cope with agents that are not controlled by our algorithm (e.g., human-driven vehicles), the computed motion will not exactly match the reality.
This yields prediction errors for the uncontrolled agents, which then can lead to infeasible optimization problems or imminent collisions.
To account for these prediction errors, we introduce an additional safety distance for the ego agent to all uncontrolled agents as a \textit{soft constraint}.
Inspired by the formulation in \cite{Gutjahr2017}, we introduce \textit{slack variables} $\slack$ to the agent-to-agent collision constraints.
With the desired safety distance of both agents $\agentSafetyDist(\t)$ and a set of slack variables $\slack_x(\t), \slack_y(\t) \in [0, ~ \agentSafetyDist(\t)]$, we modify \refEquation{eq:rear_rear_no_collision_constaint} to
\begin{subequations}
	\begin{align}
	\x^i(\t) &\leq \x^j(\t) - R^{i+j} - \agentSafetyDist(\t) + \slack_x^{ij}(\t)+ M \carToCarCollision^{ij}_1(\t)\\ 
	\x^i(\t) &\geq \x^j(\t) + R^{i+j} + \agentSafetyDist(\t) - \slack_x^{ij}(\t) - M \carToCarCollision^{ij}_2(\t)\\
	\y^i(\t) &\leq \y^j(\t) - R^{i+j} - \agentSafetyDist(\t) + \slack_y^{ij}(\t) + M \carToCarCollision^{ij}_3(\t)\\ 
	\y^i(\t) &\geq \y^j(\t) + R^{i+j} + \agentSafetyDist(\t) - \slack_y^{ij}(\t) - M \carToCarCollision^{ij}_4(\t)\\
	3 &\geq \sum_{a=1}^4  \carToCarCollision^{ij}_a(\t) ~ \forall ~ \t \in \timeintervall. 
	\end{align}
	\label{eq:rear_rear_no_collision_constaint_slack}
\end{subequations}
The slack variables will be included in the cost function (see \refSection{sec:costfunction}).
The optimizer will then seek to keep the slack variables as small as possible. Consequently, in \refEquation{eq:rear_rear_no_collision_constaint_slack}, the additional safety distance $\agentSafetyDist$ will be as high as possible. 
With this concept, fatal prediction errors (immanent collisions) are mitigated.
Note that adding a hard safety margin only leads to more conservative behavior and does not avoid infeasible optimization problems. 

To prevent collisions between the rear part of agent $\Agent^i$ and the front part of agent $\Agent^j$, we again formulate logical constraints that a collision occurs at $\t$ if and only if
\begin{align}
\x^i(\t) \geq \xfrontlb^j(\t) - R^{i+j} &\land \x^i(\t) \leq \xfrontub^j(\t) + R^{i+j} \nonumber\\
\land\ \y^i(\t) \geq \yfrontlb^j(\t) - R^{i+j} &\land \y^i(\t) \leq \xfrontub^j(\t) + R^{i+j}.
\label{eq:front_rear_collision}
\end{align}
Here we force the point $(\x^i,~ \y^i)$ to be outside the front axle approximation rectangle enlarged by the sum of the collision circle radii.
The set of constraints is formulated as described for the rear part-to-rear part collision case.
Another analog set of constraints prevents collisions between the rear part of $\Agent^j$ and the front part of $\Agent^i$.

We avoid collisions between the fronts of agents $\Agent^i$ and $\Agent^j$ applying the same strategy but forcing the center point of the front axle approximation rectangle of $\Agent^j$ to retain sufficient distance to the front axle approximation rectangle of $\Agent^i$.
Concretely, we define the sufficient distance as $R^{i+j}$ plus the size of the approximation rectangle of agent $\Agent^j$.
Hence, no front part-to-front part collision occurs at time step $\t$ if and only if
\begin{align}
\frac{1}{2} \big(\xfrontub^j(\t)+\xfrontlb^j(\t)\big) &\leq \xfrontlb^i(\t) - R^{i+j} - \frac{1}{2} \big(\xfrontub^j(\t)-\xfrontlb^j(\t)\big) \lor \nonumber\\
\frac{1}{2} \big(\xfrontub^j(\t)+\xfrontlb^j(\t)\big) &\geq \xfrontub^i(\t) + R^{i+j} + \frac{1}{2} \big(\xfrontub^j(\t)-\xfrontlb^j(\t)\big) \lor \nonumber\\
\frac{1}{2} \big(\yfrontub^j(\t)+\yfrontlb^j(\t)\big) &\leq \yfrontlb^i(\t) - R^{i+j} - \frac{1}{2} \big(\yfrontub^j(\t)-\yfrontlb^j(\t)\big) \lor \nonumber\\
\frac{1}{2} \big(\yfrontub^j(\t)+\yfrontlb^j(\t)\big) &\geq \yfrontub^i(\t) + R^{i+j} + \frac{1}{2} \big(\yfrontub^j(\t)-\yfrontlb^j(\t)\big).
\label{eq:front_front_no_collision}
\end{align}
From \refEquation{eq:front_front_no_collision} we again derive a set of constraints as in the rear part-to-rear part collision case.

As an alternative, we also formulated tighter front part-to-front part collision constraints by excluding the interval overlap of $[\xfrontlb^i - R^i, \xfrontub^i + R^i]$ with $[\xfrontlb^j - R^j , \xfrontub^j + R^j ]$ and in $y$-direction vice versa.
This formulation is slower as it needs more binary decision variables and does not provide huge benefits. 
In dense and coupled scenarios, \refEquation{eq:front_front_no_collision} can lead to harsh braking or acceleration in front of narrow but still drivable passages due to the over-conservative approximation.

\subsection{Joint Cost Function}
\label{sec:costfunction}
As the objective of the optimization problem, we chose to minimize a weighted sum of individual cost functions per agent.
This approach if referred to centralized planning in a model predictive control setting.
We formulate the individual cost function term of agent $\Agent^i$ as 
\begin{align}
J^i = \sum_{\t \in \timeintervall} \bigg( &q^i_{p} \big(\x^i(\t) - \xref^i(\t) \big)^2 + q^i_{p} \big(\y^i(\t) - \yref^i(\t) \big)^2 \nonumber \\
  + ~& q^i_{u} \ux^i(\t)^2 + q^i_{u} \uy^i(\t)^2\bigg)
  \label{eq:objective_i}
\end{align}
with suitable weighting factors $q_\square$ consisting of terms for tracking the reference trajectory and terms for penalizing the jerk.

To leverage the interests of the agents, we introduce \textit{scaling factors} $\lambda \in [0,1]$ in the overall cost function.
With these, we can push the optimization problem to generate egoistic, symmetric, or altruistic solutions.
For each agent $\Agent^i$, we define a scaling factor $\lambda^i$ with the following properties. First, 
\begin{align}
\sum_{i \in \setOfAgents} \lambda^i = 1.
  \label{eq:lambda_sum}
\end{align}
The scaling factor of the ego agent $\lambda^{\text{ego}}$ is chosen freely within the interval $[0,1]$. 
We then set
\begin{align}
\lambda^i = \frac{1-\lambda^{\text{ego}}}{|\setOfAgents|-1} ~ \forall ~ i \neq \text{ego}.
  \label{eq:lambda_other}
\end{align}
The intuitive explanation of \refEquation{eq:lambda_other} is that high values for $\lambda^{\text{ego}}$ will lead to egoistic ego behavior, all values equal a symmetric solution, and for small values of $\lambda^{\text{ego}}$ the ego agent will not enforce its own goals, only trying to fulfill the constraints.

The overall cost function is then defined as
\begin{align}
J = \sum_{i \in \setOfAgents} \lambda^i J^i  + q_\slack \sum_{\t \in \timeintervall} \sum_{i \in \setOfAgents,\ j \in \setOfAgents \setminus i} \big( \slack^{ij}_x(\t) + \slack^{ij}_y(\t) \big)^2
  \label{eq:objective}
\end{align}
with the second term penalizing high values of the slack variables and a weighting factor $q_\slack$.

\begin{figure*}[t]
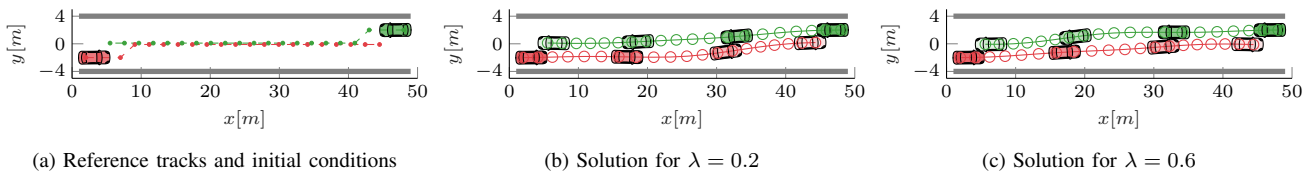

	\vspace{0.15cm}
	\begin{center}
		\begin{subfigure}{0.32\textwidth}
			\centering
			\scriptsize
			\newcommand\figurescale{0.4}
			\input{Evaluations/test_negotiation_scenario_overlapping_reference_2020-02-25__11-02-34-879/negotiation_scenario_reference.tex}
			\caption{Reference tracks and initial conditions}
			\label{fig:eval_negotiation_reference}
		\end{subfigure}
		\begin{subfigure}{0.32\textwidth}
			\centering
			\scriptsize 
			\newcommand\figurescale{0.4}
			\input{Evaluations/test_negotiation_scenario_overlapping_reference_2020-02-25__11-02-34-879/negotiation_scenario_traj_0.2.tex}
			\caption{Solution for $\lambda = 0.2$}
			\label{fig:eval_negotiation_1}
		\end{subfigure}
		\begin{subfigure}{0.32\textwidth}
			\centering
			\scriptsize
			\newcommand\figurescale{0.4}
			\input{Evaluations/test_negotiation_scenario_overlapping_reference_2020-02-25__11-02-34-879/negotiation_scenario_traj_0.6.tex}
			\caption{Solution for $\lambda = 0.6$}
			\label{fig:eval_negotiation_2}
		\end{subfigure}
		\caption{A scenario with conflicting references. The optimized solution is leveraged with the cooperation factor $\lambda$.}
		\label{fig:eval_negotiation}
	\end{center}
	\vspace{-0.15cm}
\end{figure*}

\subsection{Receding Horizon Formulation of the MIQP}
We execute the algorithm in a receding horizon fashion and, therefore, only execute the first step of the optimized trajectory.
The instance of the MIQP model solved per time step is formulated as in our previous work \cite{Esterle2020} sketched in \refSection{sec:preliminaries}, the constraints here are calculated individually for each agent.
The model is enhanced with the agent-to-agent collision avoidance constraints introduced in \refSection{sec:agenttoagentcollision}. We minimize the joint cost function \refEquation{eq:objective} defined in \refSection{sec:costfunction}.
As initial conditions, we set the states of each agent and the current region of each agent. Adding the latter helps to avoid infeasible problems on the region boundaries. In such cases, where the initial region is possibly ambiguous due to numerical inaccuracies, we start the optimization for both possible regions.
Due to the separation of $x$- and $y$-directions, absolute terms such as a reference path or the velocity along this path, cannot be included directly in the cost function. 
As we still consider it desirable to track a reference speed along a reference path, we compute the trajectory reference, a sequence of $x$ and $y$ coordinates along the discrete time $\t$, from path and speed reference values in a preprocessing step.

To speed up the solution computation and to enforce finding consistent solutions close to the solution of the previous step, we warm-start the MIQP solver using the solution from the previous receding horizon instance \cite{Marcucci2019}.
From this previous solution, we use the decision variables from step two on and perform a simple extrapolation to initialize the variables at the last time step which introduces only a neglectable overhead regarding computation time.
We leave a more advanced strategy, such as warm-starting the cuts or the branch-and-bound tree, to future work.

\section{Evaluation}
We demonstrate both the implications of the introduced cooperation factor $\lambda$ in the joint cost function and the agent-to-agent collision constraint including the soft constraint term to account for model inaccuracies in two simulated scenarios.
The algorithms are written in Mathworks MATLAB and the generated MIQP is solved using IBM Cplex.

\subsection{Levels of Cooperation in a Negotiation Scenario}
We evaluate the effect of the cooperation factor $\lambda$ in a negotiation scenario with two agents $\Agent^1$ and $\Agent^2$.
The scenario is fully symmetric with two vehicles placed on a two-lane road in oncoming direction with road boundaries. 
Both reference lines do not track the lane centers but the road center. 
This yields conflicting goals (\refFigure{fig:eval_negotiation_reference}). 
The solution can be balanced to favor one or the other agent. 
The vehicle trajectories change as $\lambda$ changes (\refFigure{fig:eval_negotiation_1} and \refFigure{fig:eval_negotiation_2}).

In \refTable{tab:eval_negotiation}, we qualitatively show the effect of varying $\lambda$.
We analyze a single run of the algorithm.
By Dist. $\Agent^\square$ we denote the accumulated distance over all $N$ time steps in meters from the solution trajectory to the reference for the respective agent.
The column Idx $\Agent^\square$ indicates at which time index the respective agent has reached the reference trajectory.
We also state the contribution of each agent to the global cost function, denoted by Cost  $\Agent^\square$.
All three metrics show the same trend; by varying $\lambda$ the respective agent is favored.
We observe that for $\lambda \approx 0.5$ all metrics are balanced, but also with a strong favor of one agent still, valid solutions are computed.

\begin{table}[tb]
\caption{Quantitative evaluation of the negotiation scenario. We compare the overall distance to the reference, the time the reference is reached, and the contributions to the cost function of each agent.}
\begin{center}
\scriptsize
\begin{tabular}{l|llllll}
$\lambda$ & Dist. $\Agent^1$ & Dist. $\Agent^2$ & Idx $\Agent^1$ & Idx $\Agent^2$ & Cost $\Agent^1$ & Cost $\Agent^2$ \\
\hline
0 & 54.684 & 19.645 & - & 11 & 3.7452 & 5006.2 \\
0.1 & 33.872 & 20.181 & 18 & 11 & 899.19 & 4568.5 \\
0.2 & 30.788 & 21.362 & 17 & 12 & 1552.6 & 4183.6 \\
0.3 & 28.937 & 22.543 & 17 & 14 & 2114.1 & 3796.1 \\
0.4 & 28.383 & 22.894 & 17 & 14 & 2731.3 & 3305.4 \\
0.5 & 22.98 & 28.297 & 14 & 17 & 2773 & 3380.1 \\
0.6 & 22.884 & 28.383 & 14 & 17 & 3304.4 & 2731.3 \\
0.7 & 22.532 & 28.929 & 14 & 17 & 3795.5 & 2113.2 \\
0.8 & 21.364 & 30.772 & 12 & 17 & 4183.5 & 1551.5 \\
0.9 & 20.168 & 33.865 & 11 & 18 & 4568.3 & 898.88 \\
1 & 19.646 & 54.666 & 11 & - & 5006.2 & 3.699 \\
\end{tabular}%

\end{center}
\label{tab:eval_negotiation}
\end{table}%

\begin{figure}[b]
	\begin{center}
		\scriptsize
		\newcommand\figurescale{0.48}
		\input{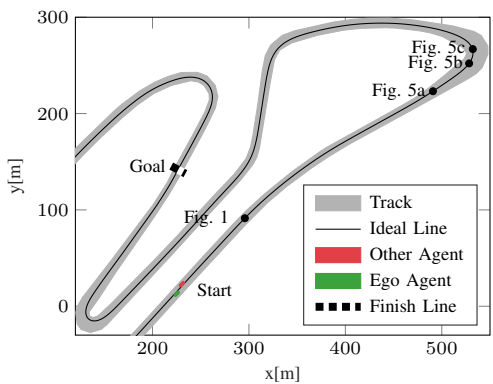}
		\caption{Setup of the racing scenario. The start points of scenes depicted in the other figures are located on the track.}
		\label{fig:initial_conditions_racetrack}
	\end{center}
	\vspace{-0.15cm}
\end{figure}

\begin{figure*}[t]
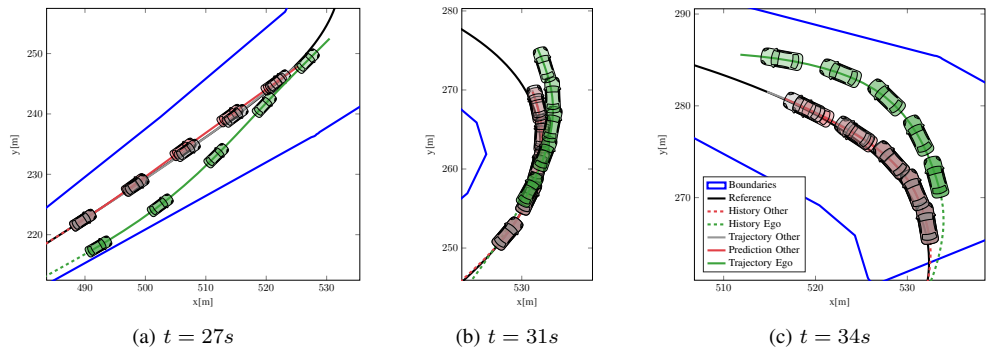

	\vspace{0.15cm}
	\begin{center}
		\begin{subfigure}[b]{0.264\textwidth}
			\includegraphics[width=1\linewidth]{Evaluations/Racetrack/racetrack_lambda_73_k_135.tikz}
			\caption{$t=27s$}
			\label{fig:overtake_135}
		\end{subfigure}
		\qquad
		\begin{subfigure}[b]{0.1275\textwidth}
			\includegraphics[width=1\linewidth]{Evaluations/Racetrack/racetrack_lambda_73_k_155.tikz}
			\caption{$t=31s$}
			\label{fig:overtake_155}
		\end{subfigure}
		\qquad
		\begin{subfigure}[b]{0.247\textwidth}
			\includegraphics[width=1\linewidth]{Evaluations/Racetrack/racetrack_lambda_73_k_170.tikz}
			\caption{$t=34s$}
			\label{fig:overtake_170}
		\end{subfigure}
		\caption{The ego agent $\Agent^1$ overtaking in the first curve with cooperation factor $\lambda = 0.3$. \refFigure{fig:intro} depicts $t=9s$.}
		\label{fig:racing_overtake}
	\end{center}
	\vspace{-0.15cm}
\end{figure*}

\subsection{Competitive Racing}
We evaluate our algorithm in an autonomous vehicle racing scenario, in which dense interactions with other agents usually occur. The goals of each agent are conflicting as each agent wants to win the race.
Besides the observations of other agents no communication is involved.
The uncertainty in the intent of other agents is neglectable compared with road traffic in this setting.
We assume the optimal line on the racetrack to be known and equal for all agents.
Therefore, each agent competes to stay on this ideal line, which we use as a reference path.
In the curves, we limit the maximum velocity with respect to the maximum lateral acceleration possible for the vehicle model.

We use a racing track of Berlin
by \citet{Heilmeier2019} with the provided ideal line and track boundaries (\refFigure{fig:initial_conditions_racetrack}). 
We naively triangulate the track boundaries and greedily merge the triangles to get the convex approximation of the environment. 
We define the finish line after 1177 m.
The ego agent $\Agent^1$ starts with a disadvantage but has a slightly higher top speed than the other agent $\Agent^2$, so it can eventually overtake and win the race.
The ego agent $\Agent^1$ uses our proposed multi-agent MIQP method to plan its behavior and trajectory, whereas the other agent $\Agent^2$ only tracks the reference line with respect to its kinematic constraints without considering the ego agent.
\refTable{tab:parameters} states the scenario parameters.
\begin{table}[b]
	\caption{Parameter values of the racing scenario.}
	\centering
	\scriptsize
	\begin{tabular}{cc|ccc|c|cc}
		$v_{des.}^1$  &
		$v_{des.}^2$  & 
		$q_{p}^i$ & 
		$q_{u}^i$ &
		$q_{\slack}$ & 
		$D$ &
		$\deltat$ & 
		$N$ \\ 
		\hline 
		$12 m/s$ 
		& $14 m/s$
		& $10 \lambda^i$ 
		& $0.5 \lambda^i$\
		& $30$ 
		& $3m$
		& $0.2s$ 
		& 20
	\end{tabular} 
	\label{tab:parameters}
\end{table}

In the overtaking scene in \refFigure{fig:intro} and \refFigure{fig:racing_overtake}, we show exemplary for $\lambda = 0.3$ how $\Agent^1$ finally makes use of its higher top speed.
We observe, that even with a very inaccurate model of $\Agent^2$ (\refFigure{fig:intro}), $\Agent^1$ can safely catch up (\refFigure{fig:overtake_135}), finally overtake (\refFigure{fig:overtake_155}), and keep the leading position even though it has to take a wider curve due to its higher speed (\refFigure{fig:overtake_170}). 

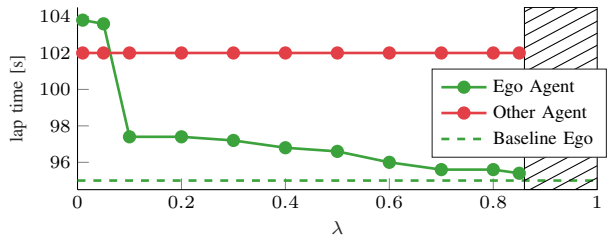
\begin{figure}[t]
	\begin{center}
		\scriptsize
		\setlength{\figH}{2.5cm}
		\setlength{\figW}{0.8\columnwidth}
		\centering
		\definecolor{mycolor1}{rgb}{0.88627,0.24706,0.27451}%
\definecolor{mycolor2}{rgb}{0.23529,0.63529,0.23529}%
\definecolor{mycolor3}{rgb}{0.58824,0.58824,0.58824}%
\begin{tikzpicture}

\begin{axis}[%
width=\figW,
height=0.97\figH,
at={(0\figW,0\figH)},
scale only axis,
unbounded coords=jump,
xmin=0,
xmax=1,
xlabel style={font=\color{white!15!black}},
xlabel={$\lambda$},
ymin=94.5,
ymax=104.5,
ylabel style={font=\color{white!15!black}},
ylabel={lap time [s]},
axis background/.style={fill=white},
axis x line*=bottom,
axis y line*=left,
legend style={at={(0.85,0.175)}, anchor=south, legend columns=1, legend cell align=left, align=left, draw=white!15!black}
]
\addplot [color=mycolor2, line width=1.0pt, mark=*, mark options={solid, mycolor2}]
  table[row sep=crcr]{%
0.01	103.8\\
0.05	103.6\\
0.1	97.4\\
0.2	97.4\\
0.3	97.2\\
0.4	96.8\\
0.5	96.6\\
0.6	96\\
0.7	95.6\\
0.8	95.6\\
0.85	95.4\\
0.9	nan\\
};
\addlegendentry{Ego Agent}

\addplot [color=mycolor1, line width=1.0pt, mark=*, mark options={solid, mycolor1}]
  table[row sep=crcr]{%
0.01	102\\
0.05	102\\
0.1	102\\
0.2	102\\
0.3	102\\
0.4	102\\
0.5	102\\
0.6	102\\
0.7	102\\
0.8	102\\
0.85	102\\
0.9	nan\\
};
\addlegendentry{Other Agent}

\addplot [color=mycolor2, dashed, line width=1.0pt]
  table[row sep=crcr]{%
0	95\\
1	95\\
};
\addlegendentry{Baseline Ego}

\addplot[area legend, draw=black, forget plot]
table[row sep=crcr] {%
x	y\\
0.86	94.5\\
1	94.5\\
1	104.5\\
0.86	104.5\\
}--cycle;
\addplot [color=black, forget plot]
  table[row sep=crcr]{%
0.957228723090988	94.5\\
1	95.1507531701494\\
nan	nan\\
0.913781301819925	94.5\\
1	95.8117936901018\\
nan	nan\\
0.870333880548863	94.5\\
1	96.4728342100541\\
nan	nan\\
0.86	95.0038133803132\\
1	97.1338747300065\\
nan	nan\\
0.86	95.6648539002656\\
1	97.7949152499588\\
nan	nan\\
0.86	96.325894420218\\
1	98.4559557699112\\
nan	nan\\
0.86	96.9869349401703\\
1	99.1169962898636\\
nan	nan\\
0.86	97.6479754601227\\
1	99.7780368098159\\
nan	nan\\
0.86	98.3090159800751\\
1	100.439077329768\\
nan	nan\\
0.86	98.9700565000274\\
1	101.100117849721\\
nan	nan\\
0.86	99.6310970199798\\
1	101.761158369673\\
nan	nan\\
0.86	100.292137539932\\
1	102.422198889625\\
nan	nan\\
0.86	100.953178059885\\
1	103.083239409578\\
nan	nan\\
0.86	101.614218579837\\
1	103.74427992953\\
nan	nan\\
0.86	102.275259099789\\
1	104.405320449483\\
nan	nan\\
0.86	102.936299619742\\
0.962775468541175	104.5\\
nan	nan\\
0.86	103.597340139694\\
0.919328047270112	104.5\\
nan	nan\\
0.86	104.258380659646\\
0.875880625999049	104.5\\
nan	nan\\
};
\end{axis}
\end{tikzpicture}%
		\caption{How can the ego agent $\Agent^1$ win the race? $\Agent^1$ incorporates the motion of $\Agent^2$. It looses the race if parameterized too conservative ($\lambda < 0.1$) and overtakes the sooner the more aggressive $\lambda$ is chosen. If the interests of $\Agent^2$ are ignored, eventually a crash occurs while overtaking (hatched area, $\lambda > 0.85$).}
		\label{fig:eval_laptime}
	\end{center}
\end{figure}
In \refFigure{fig:eval_laptime}, we analyze how different scaling factors $\lambda$ lead to different agent behavior by tracking the lap time of both agents.
As a baseline, we simulate each agent alone on the racetrack.
If $\Agent^1$ ignores the predicted motion and intention of $\Agent^2$ ($\lambda>0.85$), it drives too aggressive and eventually provokes a crash of both vehicles.
In the other extreme ($\lambda<0.1$), $\Agent^1$ ignores its own goal, which leads to very passive behavior. 
It cannot successfully overtake even though it could accelerate to higher top speed.
For scaling factors in between, we observe that depending on the $\lambda$,\ $\Agent^1$ sooner or later successfully overtakes and wins the race.
Hence, with a balanced scaling factor $\Agent^1$ can achieve its own goal to drive at higher top speed, also taking into account the intent of $\Agent^2$ to stay on the ideal line at a lower speed.

\begin{figure*}[tb]
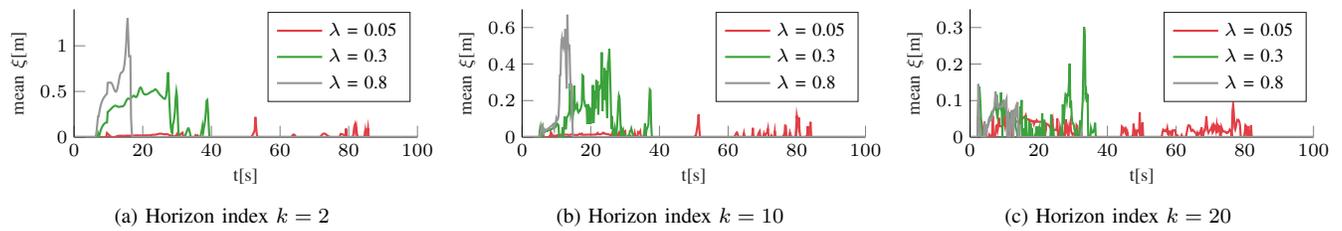

	\vspace{0.15cm}
	\setlength{\figH}{1.7cm}
	\setlength{\figW}{0.27\textwidth}
	\begin{subfigure}{0.33\textwidth}
		\centering	
		\scriptsize
		\input{Evaluations/slack_dist_2.tex}
		\subcaption{Horizon index $\t=2$}
		\label{fig:slack_eval_prediction_1}
	\end{subfigure}
	\begin{subfigure}{0.33\textwidth}
		\centering	
		\scriptsize
		\input{Evaluations/slack_dist_10.tex}
		\subcaption{Horizon index $\t=10$}
	\end{subfigure}
	\begin{subfigure}{0.33\textwidth}
		\centering	
		\scriptsize
		\input{Evaluations/slack_dist_20.tex}
		\subcaption{Horizon index $\t=20$}
		\label{fig:slack_eval_prediction_3}
	\end{subfigure}
	\caption{The effect of soft constraints coping with prediction errors. At the respective horizon index, we depict the mean safety distance compensated by the slack variables $\xi$ for different factors $\lambda$.}
	\label{fig:eval_slack_prediction}
	\vspace{-0.15cm}
\end{figure*}
In \refFigure{fig:eval_slack_prediction}, we analyze the effect of the scaling factor $\lambda$ on the interaction of the agents and the safety distance the ego agent $\Agent^1$ is willing to keep.
If $\Agent^1$ behaves aggressively, it implicitly models, that $\Agent^2$ will make room, which it will not.
This results in a high prediction error for high values of $\lambda$ which is compensated by the slack terms. 
We observe that more slack is used to mitigate immanent collisions at the beginning of the planning horizon (\refFigure{fig:slack_eval_prediction_1}).
At the end of the horizon, the magnitude of slack used is lower as the optimizer has cheaper alternatives to avoid collisions (changing position or jerk) even though the absolute errors from a non-ideal prediction is higher (\refFigure{fig:slack_eval_prediction_3}).
We show three different values of $\lambda$ representing three different behaviors of $\Agent^1$; early overtaking ($\lambda=0.8$, until $t=17s$), late overtaking ($\lambda=0.3$, until $t=40s$, also see \refFigure{fig:racing_overtake}), and no overtaking ($\lambda=0.05$).
As soon as the overtaking maneuver has been performed successfully, the slack costs tend to zero, as the safety distance can trivially be fulfilled. 
In the edge case $\lambda=0.05$, $\Agent^1$ behaves very passive and always tries to keep the safety distance big. 
In the other extreme of $\lambda=0.9$, $\Agent^1$ behaves too aggressive, accepts a high prediction error that cannot be compensated by the slack terms and finally provokes a crash.
For $\lambda=0.8$, $\Agent^1$ performs an aggressive, but still safe, overtaking maneuver, whereas for $\lambda=0.3$ it initiates the overtaking late as more drivable area is available.
Hence, with appropriate parameter selection, our algorithm can cope with not completely known cost functions of other agents and the driving style of the ego agent, including its willingness to take risks can be adopted.

\section{Conclusion and Future Work}
\label{sec:conclusion}
We introduced a novel linear differential game formulation for the multi-agent behavior planning problem leveraging the interests of all agents. 
Using linear constraints for checking collision between agents, we formulate a mixed-integer quadratic program and obtain an optimal solution. 
The introduction of slack variables to the collision constraints makes our method robust against inaccuracies of the modeled future motion of other agents.

We first studied the impact of an altruistic, cooperative, and egoistic behavior setting on the outcome of a symmetric scenario with conflicting goals of both agents.
We then demonstrated the effectiveness of our method in a competitive autonomous vehicle racing scenario, where we introduced inaccuracies between the true and modeled behavior of the other agent.
In this highly interactive scenario that poses high requirements on the receding horizon replanning scheme, we showed that the proposed multi-agent planning approach correctly satisfies all model constraints and also copes with inaccurate agent models.
Furthermore, the introduced cooperation factor in the joint cost function allows leveraging the goals of either agent.
In combination with our previous work \cite{Esterle2020}, which proved correctness in the model of the vehicle kinematics and the obstacle avoidance, we have successfully demonstrated the feasibility in simulation.

As a next step, we plan to apply the formulation in real-road driving scenarios using the institute’s research vehicle \cite{Kessler2019a}. 
For this to be feasible regarding computation time, we need to find a trade-off which agents to denote as interacting and which others to predict maneuvers for solely. 
The approach is applicable for an arbitrary number of agents, but an analysis on the computational scalability has to be conducted. 
Also, the assumption on perfect observation of other agents has to be weakened by introducing suitable observers and only optimizing within the sensing radius.
Uncertainty measures from the perception pipeline shall be used to parametrize the soft constraints online.
Furthermore, modeling traffic rules as logical constraints, which pose constraints to the possible behavior to choose from, shall be elaborated to better approximate future behaviors.

\section*{Acknowledgment}
This research was partly funded by the Bavarian Ministry of Economic Affairs, Regional Development and Energy, project Dependable AI and supported by the Intel Collaborative Research Institute - Safe Automated Vehicles.

\renewcommand{\bibfont}{\footnotesize}
\printbibliography
\end{document}